\documentclass[10pt,a4paper]{article}
\usepackage{lrec-coling2024} 

\usepackage{enumerate}
\usepackage{graphicx, float}
\usepackage{hyperref}

\title{VNLP: Turkish NLP Package}

\name{Melikşah Türker, Mehmet Erdi Arı, Aydın Han} 
\address{VNGRS \\
         YTÜ Teknopark B2 103 Davutpaşa, İstanbul, Turkey \\
         \{meliksah.turker, erdi.ari, aydin.han\}@vngrs.com\\}
         
\abstract{
In this work, we present VNLP: the first dedicated, complete, open-source, well-documented, lightweight, production-ready, state-of-the-art Natural Language Processing (NLP) package for the Turkish language.
It contains a wide variety of tools, ranging from the simplest tasks, such as sentence splitting and text normalization,
to the more advanced ones, such as text and token classification models.
Its token classification models are based on "Context Model", a novel architecture that is both an encoder and an auto-regressive model.
NLP tasks solved by VNLP models include but are not limited to Sentiment Analysis, Named Entity Recognition, Morphological Analysis \& Disambiguation and Part-of-Speech Tagging.
Moreover, it comes with pre-trained word embeddings and corresponding SentencePiece Unigram tokenizers.
VNLP has an open-source GitHub repository,
ReadtheDocs documentation, PyPi package for convenient installation, Python and command-line API and a demo page to test all the functionality.
Consequently, our main contribution is a complete, compact, easy-to-install and easy-to-use NLP package for Turkish.
\\ \newline 
}

\begin{document}

\maketitleabstract

\section{Introduction} 
\label{sec:intro}
Although frequently considered a low-resource language, Turkish Natural Language Processing (NLP) research has recently attracted more attention~\cite{rw_mukayese_safaya2022mukayese, rw_turkishdelight_alecakir2022turkishdelightnlp, rw_baykara_seq2seq_baykara2022turkish, rw_turkishnlpresources_ccoltekin2023resources,rw_boun_tulap_uskudarli2023tulap}.
Despite this increased attention, there is a gap between research papers and their inference-ready tools.
In most cases, the research paper is specific to one or a few NLP tasks with a GitHub repository that allows reproducibility of evaluation metrics and contains open-source codes.
However, this is nowhere near a complete, state-of-the-art, well-documented, lightweight and inference-ready tool.

Seeing this gap, we present VNLP to be the solution.
VNLP contains a wide range of tools, namely;
Sentence Splitter, Text Normalizer, Named Entity Recognizer, Part-of-Speech Tagger, Dependency Parser, Morphological Analyzer \& Disambiguator and Sentiment Analyzer.
Deep Learning models in VNLP are very compact and lightweight, ranging from 2.3M to 5.7M parameters.
Moreover, we release the pre-trained word embeddings and corresponding SentencePiece Unigram tokenizers that are used by these models.

The code base is well-structured, readable and comes with documentation hosted on ReadtheDocs.
\footnote{\href{https://vnlp.readthedocs.io/en/latest/}{https://VNLP.readthedocs.io/en/latest/}}.
The package is available on PyPi
\footnote{\href{https://pypi.org/project/vngrs-nlp/}{https://pypi.org/project/vngrs-nlp/}}
and can be installed using pip.
\footnote{\texttt{pip install vngrs-nlp}}.
It has Python and CLI APIs that allow integration with other systems.
Lastly, there is a demo page
\footnote{\href{https://demo.vnlp.io/}{https://demo.VNLP.io/}}
where the mentioned models can be tested.

\begin{figure*}[]
\includegraphics[width=2.0\columnwidth]{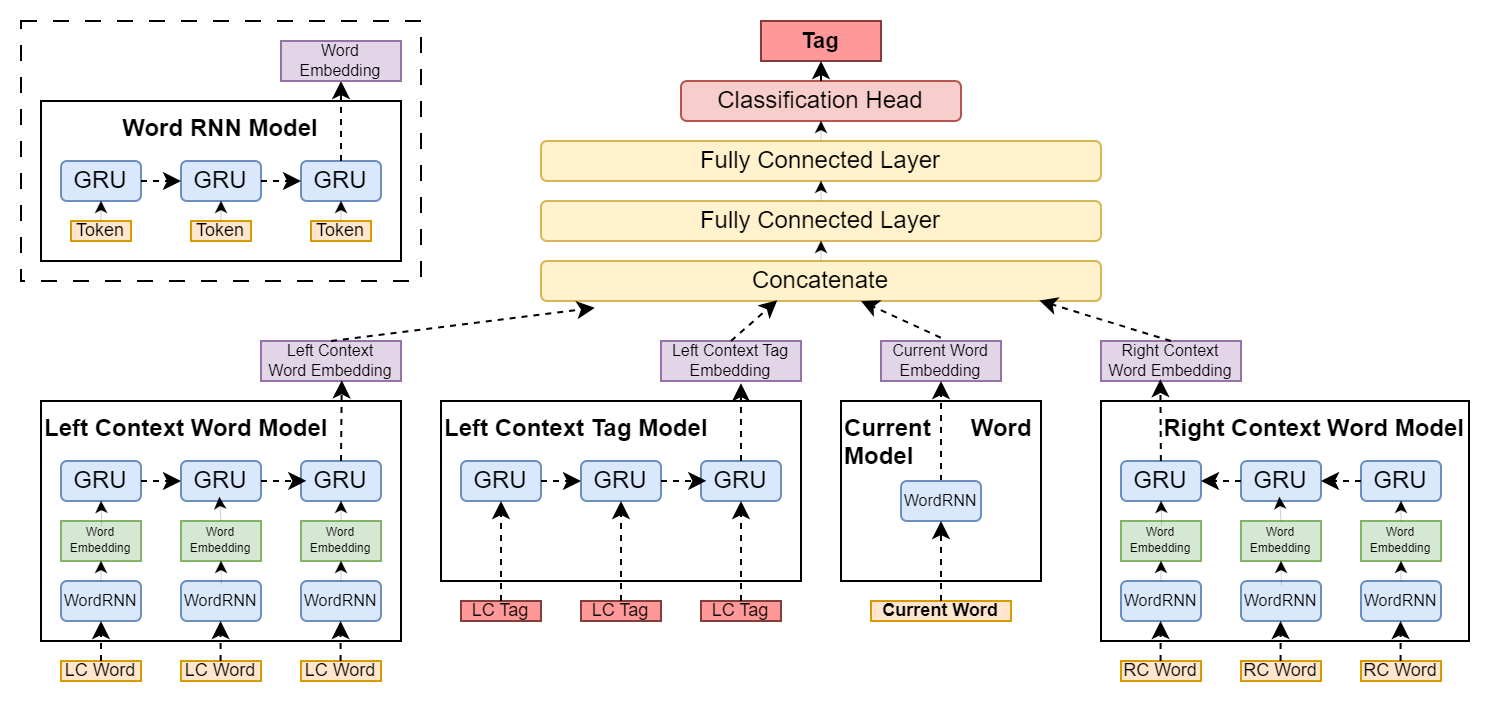}
\caption{Context Model consists of 4 components. Word Model that processes the subword tokens in a word, Left Context Model that processes words in the left context left-to-right, Right Context Model that processes words in the right context right-to-left and Left Context Tag Model that processes the classified tags in the left context.}
\label{fig:context_model}
\end{figure*}

\section{Related Work}
NLP research has attracted a significant amount of research in the past decade.
The development and publication of text-processing technologies have taken a crucial role.
Although it was developed way back in 1997, democratization of Long Short-Term Memory(LSTM)~\cite{lstm_hochreiter1997long} via open-source Deep Learning frameworks~\cite{keras_chollet2015keras, tensorflow_tensorflow2015, pytorch_NEURIPS2019_9015} has made an important contribution in surge of NLP research by lowering the barrier to entry.
Following the ideas proposed in LSTM, the invention of Gated Recurrent Unit (GRU) allowed the reduction of the number of parameters by about 25\%, decreasing the computation cost without significant performance loss.

Word embedding models like Word2Vec and GloVe have laid the foundations for transfer learning and have been improved by their successor FastText.
Using these pre-trained word embeddings and transferring the knowledge obtained during their training to the downstream task's model improved the performance of NLP models.
However, the mentioned word embedding methods are not context-aware; that is, each word's embedding vector is static regardless of the words it is surrounded by in the downstream task.
ELMo has come to the rescue and further improved the transfer learning with context-aware word embeddings.
Frameworks like spaCy~\cite{spacy}, NLTK~\cite{nltk_bird2009natural} and gensim~\cite{gensim_rehurek_lrec} allowed anyone to use these new technologies by offering methods and pre-trained models that are ready to use in production environment for inference.
Huggingface~\cite{huggingface_transformers_wolf2019huggingface} and BERTurk~\cite{berturk} filled the gap for higher-level tasks like text and sentence classification and provided state-of-the-art results.

Turkish NLP researchers have utilized these methods to develop models for tasks such as Morphological Disambiguation, Syntactic Parsing, Dependency Parsing, Part-of-Speech Tagging, Named Entity Recognition and Sentiment Analysis for Turkish.
Often, they were developed as individual models that solve a specific problem and are published as separate research papers.
If one is lucky, the paper would contain a link to the GitHub repository that hosts the code to reproduce the results.
Rarely would the repository contain a Docker image or a CLI API to use the model for inference.

With the attempt to bring separate models and research directions for Turkish under a single banner, three recent works have been published.
Mukayese~\cite{rw_mukayese_safaya2022mukayese} has aimed to be the benchmark for datasets and task evaluation.
TurkishDelight~\cite{rw_turkishdelight_alecakir2022turkishdelightnlp} has aimed to bring several models together and serve them in a demo page.
TULAP~\cite{rw_boun_tulap_uskudarli2023tulap} has aimed to open source Turkish NLP resources developed at Boğaziçi University, offering a demo page, hosting datasets and Docker images to allow inference.
Although these tools solve the mentioned problems partially, there is still no complete NLP package for Turkish that is open-source, well-documented, PyPi installable and comes with an easy-to-use API.

\section{Functionality \& Models} 

\subsection{Sentence Splitter}
\label{sec:sentencesplitter}
Sentence splitting is the task of splitting a bulk text into separate sentences. Although this looks trivial, in order to obtain a well-working sentence splitter, there are exceptions that must be handled correctly, such as numbers and abbreviations.

For this task, we use the implementation of Koehn and Schroeder~\cite{sentencesplitter_koehnschroeder},
by simplifying the code for Turkish and expanding its lexicon of abbreviations.

\subsection{Normalizer}
Normalization is the task of standardizing the text input, which can be in different forms as it may come from various sources such as social media, customer feedback, and news articles.

\begin{table*}[]
\centering
\begin{tabular}{l c c }
         & \multicolumn{2}{c}{Accuracy} \\%
        \hline
	Dataset & Ambiguous Words & All Words \\
	\hline
        \hline
        TrMorph2006~\cite{trmorph2006_yuret} & 94.67 & 96.64\\
        TrMorph2018~\cite{trmorph2018_yuret} & 93.76 & 95.35\\
        \hline

\end{tabular}
\caption{Stemmer: Morphological Analyzer \& Disambiguator}
\label{tbl:stemmer}
\end{table*}

\begin{table*}[]
\centering
\begin{tabular}{l c c }
	Dataset & Accuracy & F1 Macro \\
	\hline
        \hline
        WikiAnn~\cite{ner_wikiann_pan2017cross} & 98.80 & 98.14\\
        Gungor~\cite{ner_joint_gungor2018improving} & 99.70 & 98.59\\
        TEGHub~\cite{ner_teghub} & 99.74 & 98.91\\
        \hline

\end{tabular}
\caption{Named Entity Recognizer}
\label{tbl:ner}
\end{table*}

\subsubsection{Spelling Correction}
Spelling correction is the task of detecting and then correcting misspelled and mistyped words.
VNLP uses Jamspell~\cite{github-jamspell}, a spell-checking library written in C++.
It was chosen over other libraries as it is faster and produces lower error rates.
Jamspell uses adjacent words to correct spelling errors, which is the reason behind the lower error rates compared to alternatives.

Jamspell requires a dictionary of word frequencies.
We use a custom dictionary file generated by training on a mixed corpus consisting of OPUS-100~\cite{opus_zhang2020improving}, Bilkent Turkish Writings~\cite{dataset-bilkent} and TED Talks~\cite{siarohin2021motion} datasets.
These datasets were chosen over others since they were observed to contain less noise.

\subsubsection{Deasciifier}
Deasciification is the process of converting a text input written in ASCII-only characters to its correct version in the target language.
VNLP Deasciifer converts the text written in an English keyboard to the correct version of how it would be had it been written using a Turkish keyboard.
It is directly taken from Sevinç's implementation~\cite{deasciifier_emresevinc}. 

\subsubsection{Number to Word}
Written texts may contain numbers in both numerical and written forms.
In order to standardize these, one could seek to convert numbers to written text forms.
Number to Word function implements this.

On top of these, the VNLP Normalizer class offers more trivial functions for lowercasing, punctuation removal and accent mark removal.

\begin{table*}[]
\centering
\begin{tabular}{l c c }
	Universal Dependencies 2.9~\cite{ud29} & LAS & UAS \\
	\hline
        \hline
        UD\_Turkish-Atis & 88.52 & 91.54\\
        UD\_Turkish-BOUN & 67.64 & 78.15\\
        UD\_Turkish-FrameNet & 81.12 & 92.30\\
        UD\_Turkish-GB & 72.97 & 88.58\\
        UD\_Turkish-IMST & 63.32 & 76.53\\
        UD\_Turkish-Kenet & 68.80 & 83.51\\
        UD\_Turkish-Penn & 70.72 & 85.24\\
        UD\_Turkish-PUD & 61.31 & 74.77\\
        UD\_Turkish-Tourism & 90.96 & 97.31\\
        \hline

\end{tabular}
\caption{Dependency Parser}
\label{tbl:dep}
\end{table*}

\subsection{Stopword Remover}
Stopwords are the words that take place in virtually any text and provide no context information alone, such as "and", "such", or "if".
While working on a wide variety of NLP tasks, one can seek to get rid of them before further analysis.
VNLP offers two algorithms to get rid of Turkish stopwords.

\subsubsection{Static Method}
The static method is the conventional method that contains a pre-defined static stopword lexicon and removes these words from the text input.
VNLP uses an improved version of the stopword lexicon offered in the Zemberek package~\cite{zemberek}.

\subsubsection{Dynamic Method}
The dynamic method is the more advanced version, where stopwords are determined depending on the given corpus.
The implemented method determines the stopwords by looking at the word frequencies\cite{stopwords_saif2014stopwords}
and their breaking point~\cite{stopwords_satopaa2011finding}
to set a threshold of frequencies and consider the words above the threshold as stopwords.
Moreover, this method is language agnostic and can be used for all kinds of texts to obtain the most frequent words.

\subsection{SentencePiece Unigram Tokenizer}
\label{sec:sentencepieceunigram}
Deep Learning models in VNLP use subword tokens, tokenized by SentencePiece Unigram Model~\cite{sentencepiece_kudo2018sentencepiece} to process the text.
SentencePiece Unigram Tokenizer is trained from scratch on a corpus of 10 GB Turkish text, which consists of random samples from OSCAR~\cite{oscar_2022arXiv220106642A}, OPUS~\cite{opus_zhang2020improving} and Wikipedia dump datasets.
It comes in 2 sizes, with vocabulary sizes of 16,000 and 32,000.

\subsection{Pre-trained Word Embeddings}
VNLP offers pre-trained word embeddings for two types of tokens.
\begin{enumerate}
    \item \textbf{TreebankWord} embeddings are tokenized by NLTK's~\cite{nltk_bird2009natural} TreebankWord Tokenizer.
    \item \textbf{SentencePiece Unigram} embeddings are tokenized by SentencePiece Unigram Tokenizer.
\end{enumerate}

Embeddings for these tokens are trained using Word2Vec~\cite{word2vec_mikolov2013efficient} and FastText~\cite{fasttext_bojanowski2017enriching} algorithms implemented by gensim~\cite{gensim_rehurek_lrec} framework. 

\subsubsection{Word2Vec}
\label{sec:pretrained_word2vec_embeddings}
Word2Vec embeddings are trained for both of the tokenization methods mentioned above.
TreebankWord tokenized Word2Vec embeddings come in 3 sizes.
\begin{itemize}
    \item \textbf{Large}: vocabulary size: 128,000, embedding dimension: 256
    \item \textbf{Medium}: vocabulary size: 64,000, embedding dimension: 128
    \item \textbf{Small}: vocabulary size: 32,000, embedding dimension: 64
\end{itemize}

SentencePiece Unigram tokenized Word2Vec embeddings come in 2 sizes:
\begin{itemize}
    \item \textbf{Large}: vocabulary size: 32,000, embedding dimension: 256
    \item \textbf{Small}: vocabulary size: 16,000, embedding dimension: 128
\end{itemize}

The difference in vocabulary sizes of the two tokenization methods is due to the fact that the Unigram tokenizer is never out of vocabulary and 32,000 is a reasonable size, being often used for state-of-the-art monolingual models~\cite{berturk, t5_2019t5}.

\subsubsection{FastText}
FastText embeddings are trained for TreebankWord tokenized tokens only.
Similar to Word2Vec configuration, they come in 3 sizes.
\begin{itemize}
    \item \textbf{Large}: vocabulary size: 128,000, embedding dimension: 256
    \item \textbf{Medium}: vocabulary size: 64,000, embedding dimension: 128
    \item \textbf{Small}: vocabulary size: 32,000, embedding dimension: 64
\end{itemize}

\begin{table*}[]
\centering
\begin{tabular}{l c c }
	Universal Dependencies 2.9~\cite{ud29} & Accuracy & F1 Macro \\
	\hline
        \hline
        UD\_Turkish-Atis & 98.74 & 98.80\\
        UD\_Turkish-BOUN & 87.08 & 78.84\\
        UD\_Turkish-FrameNet & 95.09 & 90.39\\
        UD\_Turkish-GB & 85.59 & 66.20\\
        UD\_Turkish-IMST & 90.69 & 78.45\\
        UD\_Turkish-Kenet & 91.94 & 87.66\\
        UD\_Turkish-Penn & 94.52 & 93.29\\
        UD\_Turkish-PUD & 83.87 & 65.59\\
        UD\_Turkish-Tourism & 98.45 & 93.25\\
        \hline

\end{tabular}
\caption{Part-of-Speech Tagger}
\label{tbl:pos}
\end{table*}

\begin{table*}[]
\centering
\begin{tabular}{l c c }
	~ & Accuracy & F1 Macro \\
	\hline
        \hline
        Mixture of Datasets & 94.69 & 93.81\\
        \hline
\end{tabular}
\caption{Sentiment Analyzer}
\label{tbl:sentiment}
\end{table*}

\subsection{Context Model}
\label{sec:context_model}
Context Model is the base architecture used in several deep learning models in VNLP.
It is inspired by Ref.~\cite{stemmer_shen2016role}
and consists of 4 input components as shown in Fig.~\ref{fig:context_model}.
They are Left Context Word Model, Left Context Tag Model, Current Word Model and Right Context Word Model, respectively.

The model input is tokenized by SentencePiece Unigram Tokenizer, and corresponding pre-trained Word2Vec embeddings are fed to the network.
It processes each word one by one, and each word can be represented by multiple subword tokens.
For this reason, the model contains a Word RNN Model that processes all subword tokens in a word and returns a single word embedding in the last time step.
Word RNN Model is used by Current Word, Left and Right Context Models and its parameters are shared among them.
Left and Right Context Word Models process the words on left from left to right and the words on right from right to left, respectively.
Left Context Tag Model processes the classification results of prior words.
In the end, 4 components produce 4 embedding vectors, which are concatenated and processed by 2 fully connected layers followed by a classification head that produces classification logits.

The network is made of GRU~\cite{gru_cho2014learning} cells, which provide a computation advantage over conventional LSTM~\cite{lstm_hochreiter1997long} cells.
Throughout this work, all RNNs are made of GRU cells.

The main advantage of the Context Model is it combines the idea of auto-regressive sequence-to-sequence models with token classifier encoder-only models.
This is actuated by taking the classification results of prior words on the left context as input while classifying words instead of subwords.
This schema has two benefits over BERT-based~\cite{bert_devlin2018bert} encoder-only models.
First, its auto-regressive structure allows for taking the classification results of earlier words into account.
Second, classifying words instead of tokens/subwords guarantees the alignment of words and tags.

\subsection{Stemmer: Morphological Analyzer \& Disambiguator}
Stemming is the task of obtaining the stems of the words in a sentence, depending on the context.
It is useful to standardize the text input, especially in agglutinative languages such as Turkish.

A morphological analyzer allows obtaining the stem and the morphological tags of a given word.
However, it returns multiple candidates since the word may have multiple meanings depending on the context of the sentence.
See the two examples below:

\begin{itemize}
    \item Üniversite sınavlarına canla başla çalışıyorlardı. (They were studying really hard for the university entrance exams.)
    \item Şimdi baştan başla. (Now, start over.)
\end{itemize}

In the first sentence, the word "başla" is a noun, meaning "hard", describing the struggle of studying while
in the second sentence, the word "başla" is the verb, meaning "start".

Hence, a morphological analyzer is context-agnostic and simply provides all of the potential analyses or candidates.
Given the context and the potential analyses, a morphological disambiguator selects the correct analysis result.
Stemmer class implements Shen's~\cite{stemmer_shen2016role}
morphological disambiguation model with slight differences that result from a different model config and using GRU instead of LSTM.
Stemmer consists of 2.3M parameters.
It uses Yildiz's work~\cite{yildizanalyzer} as the morphological analyzer.

Then, having a morphological disambiguator on top of a morphological analyzer allows finding the correct stem of a word depending on the context.
Stemmer model utilizes the Turkish pre-trained Word2Vec embeddings described in Section.~\ref{sec:pretrained_word2vec_embeddings} and is trained on TrMorph2006~\cite{trmorph2006_yuret}, TrMorph2016~\cite{trmorph2016_yildiz} and TrMorph2018~\cite{trmorph2018_yuret} datasets.

\subsection{Named Entity Recognizer}
Named Entity Recognition (NER) is the task of finding the named entities in a sentence.
Although there are several variants of entities and how they are represented, VNLP's Named Entity Recognizer allows finding Person, Location and Organization entities in the given sentence using IO format.
It is based on the Context Model architecture \ref{sec:context_model}, consists of 5.6M parameters and is trained on a collection of open-source NER datasets%
~\cite{ner_joint_gungor2018improving, ner_milliyetnews_tur2003statistical, ner_news_kuccuk2016named, ner_teghub, ner_turkishtweets_kuccuk2014named, ner_wikiann_pan2017cross, ner_xtreme_hu2020xtreme}.
TWNERTC~\cite{ner_twnertc_sahin2017automatically} is also considered; however, it is excluded due to being too noisy and actually deteriorating the model performance.

\begin{figure*}[ht!]
\includegraphics[width=2.0\columnwidth]{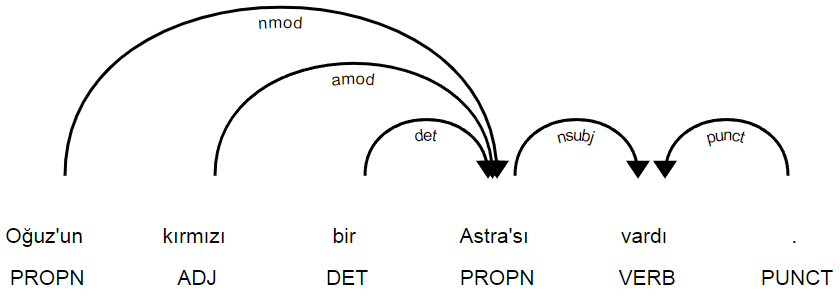}
\caption{Dependency Parser produces arcs and labels to indicate the relations between words. Part-of-speech tags below are produced by Part-of-Speech Tagger.}
\label{fig:dependency_parsing}
\end{figure*}

\subsection{Dependency Parser}
Dependency Parsing is the task of showing the dependencies of words in a sentence, along with the dependency labels.
An example can be seen in Fig.~\ref{fig:dependency_parsing}.
VNLP Dependency Parser is based on the Context Model architecture with a slight difference in classification head and left context tag inputs.
The difference arises from the fact that the model makes two classification decisions for each word, that is, arc (the index of the word it depends on) and the dependency tag.
This is implemented by a single vector in the classification head where the first part of the vector represents the arc and the second part represents the dependency tag.
Consequently, Binary Cross Entropy is used to train the model, as it is a multi-label classifier.
It consists of 5.7M parameters.
The model is trained on Universal Dependencies 2.9~\cite{ud29} dataset.

\subsection{Part-of-Speech Tagger}
Part-of-speech tagging is the task of assigning part-of-speech tags to the words in a sentence, such as nouns, pronouns, verbs, adjectives, punctuation and so on.
VNLP Part-of-Speech Tagger is also based on the Context Model architecture and consists of 2.6M parameters.

Part-of-Speech Tagger is trained on Universal Dependencies 2.9 dataset as well.

\subsection{Sentiment Analyzer}
\label{sec:sentiment}
Sentiment Analysis is the task of classifying a text into positive or negative.
Some of the typical use cases are understanding the sentiment of social media text and measuring customer satisfaction through comments and feedback.
VNLP Sentiment Analyzer implements a text classifier for this purpose.
Similar to other models in VNLP, Sentiment Analyzer uses SentencePiece Unigram Tokenized pre-trained Word2Vec embeddings and is based on GRU.
However, as it is a text classifier, compared to the word tagger models so far, its architecture is different.
It uses a stack of Bidirectional RNNs to process the input tokens, followed by a GlobalAveragePooling1D and fully connected layer before the final classification head.

Sentiment Analyzer consists of 2.8M parameters and is trained on a large mix of social media, customer comments and research datasets%
~\cite{sentiment_basturk, sentiment_bilen, sentiment_bilen2, sentiment_bounti, sentiment_coskuner, sentiment_gokmen, sentiment_guven, sentiment_ozler, sentiment_sarigil, sentiment_subasi, sentiment_kahyaoglu,  sentiment_yilmaz}.

\section{Results} \label{sec:results}
Models are trained using Adam~\cite{adamoptimizer_kingma2014adam} optimizer ($\epsilon = 1e-3$) along with a linear learning rate decay of 0.95 per epoch.
The number of epochs varies from task to task and is determined according to validation loss.

\textbf{Stemmer}: is evaluated on the test splits of TrMorph2006~\cite{trmorph2006_yuret} and TrMorph2018~\cite{trmorph2018_yuret} datasets.
A word is ambiguous if the morphological analyzer returns several candidates as the result.
Following the original work~\cite{stemmer_shen2016role}, Accuracy and F1 Macro scores are reported for both ambiguous words and all words in Table.~\ref{tbl:stemmer}.

\textbf{Named Entity Recognizer}: is evaluated on the test splits of WikiAnn~\cite{ner_wikiann_pan2017cross}, Gungor~\cite{ner_joint_gungor2018improving} and TurkishNER-BERT~\cite{ner_teghub} datasets.
Accuracy and F1 Macro scores are reported in Table.~\ref{tbl:ner}.

\textbf{Dependency Parser}: is evaluated on the test splits of Universal Dependencies 2.9~\cite{ud29} dataset.
Labelled Attachment Score (LAS) and Unlabelled Attachment Score (UAS) are reported in Table.~\ref{tbl:dep}.

\textbf{Part-of-Speech Tagger}: is evaluated on the test splits of Universal Dependencies 2.9~\cite{ud29} dataset.
Accuracy and F1 Macro scores are reported in Table.~\ref{tbl:pos}.

\textbf{Sentiment Analyzer}: is evaluated on the test split of the combined dataset mentioned in~\ref{sec:sentiment}.
It is generated by scikit-learn's~\cite{scikit-learn} train test split function with the following config: test\_size = 0.10, random\_state = 0, shuffle = True, stratify by label.
The reason we created the test split from scratch is due to the fact that the compiled dataset does not contain any pre-defined test split except for Ref.~\cite{sentiment_bounti}.
Instead of evaluating on this small sample only, we preferred evaluating on a larger and more diverse test set that comes from various sources.
Accuracy and F1 Macro scores are reported in Table.~\ref{tbl:sentiment}.

\textbf{Spelling Corrector}: is evaluated on 100 random samples taken from My Dear Watson~\cite{dataset-contradictory-my-dear-watson} and TED Talks~\cite{siarohin2021motion} datasets.
Although it scores an Accuracy of 0.69 and a Word Error Rate (WER) of 0.09 on this sample, we report these numbers for informative purposes only, as it is an under-development module of the package.
A more comprehensive study for Spelling will be conducted later on.

\section{Conclusion} \label{sec:conclusion}
We presented VNLP in this work. It is the first complete, open-source, production-ready, well-documented, PyPi installable NLP library for Turkish.
It contains a wide range of tools, including both low and high-level NLP tasks.
Implemented deep learning models are compact yet competitive.
The Context Model presented in this work brings two advantages over BERT-based classification models by taking the prediction results of earlier words into account and guaranteeing the word-tag alignments.
Hence, our contribution is a well-engineered, documented, easy-to-use NLP package based on its novel Context Model architecture.

\cleardoublepage

\section{Bibliographical References}\label{sec:reference}
\bibliography{bibliography}
\bibliographystyle{lrec-coling2024-natbib}
\end{document}